\documentclass[10pt,twocolumn,letterpaper]{article}

\usepackage[pagenumbers]{cvpr} % To force page numbers, e.g. for an arXiv version

\usepackage{mmstyle}
\usepackage{graphicx}
\usepackage{amsmath}
\usepackage{amssymb}
\usepackage{xcolor}
\definecolor{Graylight}{gray}{0.9}
\usepackage{colortbl}
\usepackage{booktabs}
\usepackage{multirow}
\usepackage{mathrsfs}
\usepackage{marvosym}
\usepackage{enumerate}
\usepackage[hang,flushmargin]{footmisc} 
\usepackage{tablefootnote}

\usepackage[accsupp]{axessibility} % Improves PDF readability for those with disabilities.

\usepackage[pagebackref,breaklinks,colorlinks]{hyperref}

\def\paperID{} % *** Enter the CVPR Paper ID here
\def\confName{CVPR}
\def\confYear{2024}

\begin{document}

%%%%%%%%% TITLE - PLEASE UPDATE

\title{InternLM-XComposer2-4KHD: A Pioneering Large Vision-Language Model Handling Resolutions from 336 Pixels to 4K HD}

\author{Xiaoyi Dong$^{*1,2}$, Pan Zhang$^{*1}$, Yuhang Zang$^{*1}$, Yuhang Cao$^{1,2}$, Bin Wang$^{1}$, Linke Ouyang$^{1}$, \\ Songyang Zhang$^{1}$, Haodong Duan$^{1}$, Wenwei Zhang$^{1}$, Yining Li$^{1}$, Hang Yan$^{1}$, Yang Gao$^{1}$, Zhe Chen$^{1}$ \\ Xinyue Zhang$^{1}$, Wei Li$^{1}$,
Jingwen Li$^{1}$,
Wenhai Wang$^{1,2}$, Kai Chen$^{1}$, Conghui He$^{3}$, Xingcheng Zhang$^{3}$, \\ Jifeng Dai$^{4,1}$, Yu Qiao$^{1}$, Dahua Lin$^{1,2}$, Jiaqi Wang$^{1,}${\textsuperscript{\Letter}}\\
$^1$Shanghai Artificial Intelligence Laboratory,  $^2$The Chinese University of Hong Kong, \\ $^3$SenseTime Group, $^4$Tsinghua University \\
\tt\small
internlm@pjlab.org.cn
}

% \twocolumn[{
% \renewcommand\twocolumn[1][]{#1}
% \maketitle
% \begin{center}
%     \centering
%     \vspace{-20pt}
%     \includegraphics[width=1.0\linewidth]{figures/cases/teaser2.pdf}
%     \setlength{\abovecaptionskip}{0mm}
%     \vspace{-10pt}
%     \captionof{figure}{\small
%         Overview of free-form text-image composition and comprehension of InternLM-XComposer2.
%         Our model based on InternLM2-7B~\cite{2023internlm} not only significantly outperforms existing multimodal models but also \textbf{matches or even surpasses GPT-4V~\cite{openai2023gpt4} and Gemini Pro~\cite{geminiteam2023gemini} in certain assessments}. (Please zoom-in to see the details.)
% 	}
% 	\label{fig:teaser}
%     \vspace{-5pt}
% \end{center}
% }]

\maketitle
% \thispagestyle{empty}
%%%%%%%% Main Text
% abstract

{\let\thefootnote\relax\footnotetext{\noindent* indicates equal contribution.}}

\begin{abstract}

The Large Vision-Language Model (LVLM) field has seen significant advancements, yet its progression has been hindered by challenges in comprehending fine-grained visual content due to limited resolution. 
Recent efforts have aimed to enhance the high-resolution understanding capabilities of LVLMs, yet they remain capped at approximately 1500 $\times$ 1500 pixels and constrained to a relatively narrow resolution range. This paper represents InternLM-XComposer2-4KHD, a groundbreaking exploration into elevating LVLM resolution capabilities up to 4K HD (3840 $\times$ 1600) and beyond. Concurrently, considering the ultra-high resolution may not be necessary in all scenarios, it supports a wide range of diverse resolutions from 336 pixels to 4K standard, significantly broadening its scope of applicability. Specifically, this research advances the patch division paradigm by introducing a novel extension: dynamic resolution with automatic patch configuration. It maintains the training image aspect ratios while automatically varying patch counts and configuring layouts based on a pre-trained Vision Transformer (ViT) (336 $\times$ 336), leading to dynamic training resolution from 336 pixels to 4K standard. Our research demonstrates that 
scaling training resolution up to 4K HD leads to consistent performance enhancements without hitting the ceiling of potential improvements. 
InternLM-XComposer2-4KHD shows superb capability that matches or even surpasses GPT-4V and Gemini Pro in 10 of the 16 benchmarks.
The InternLM-XComposer2-4KHD model series with 7B parameters are publicly available at \url{https://github.com/InternLM/InternLM-XComposer}.

\end{abstract}
% introduction
\section{Introduction}
\label{sec:intro}

\begin{figure}[t!]
    \centering
    %\vspace{-20pt}
    \includegraphics[width=1.0\linewidth]{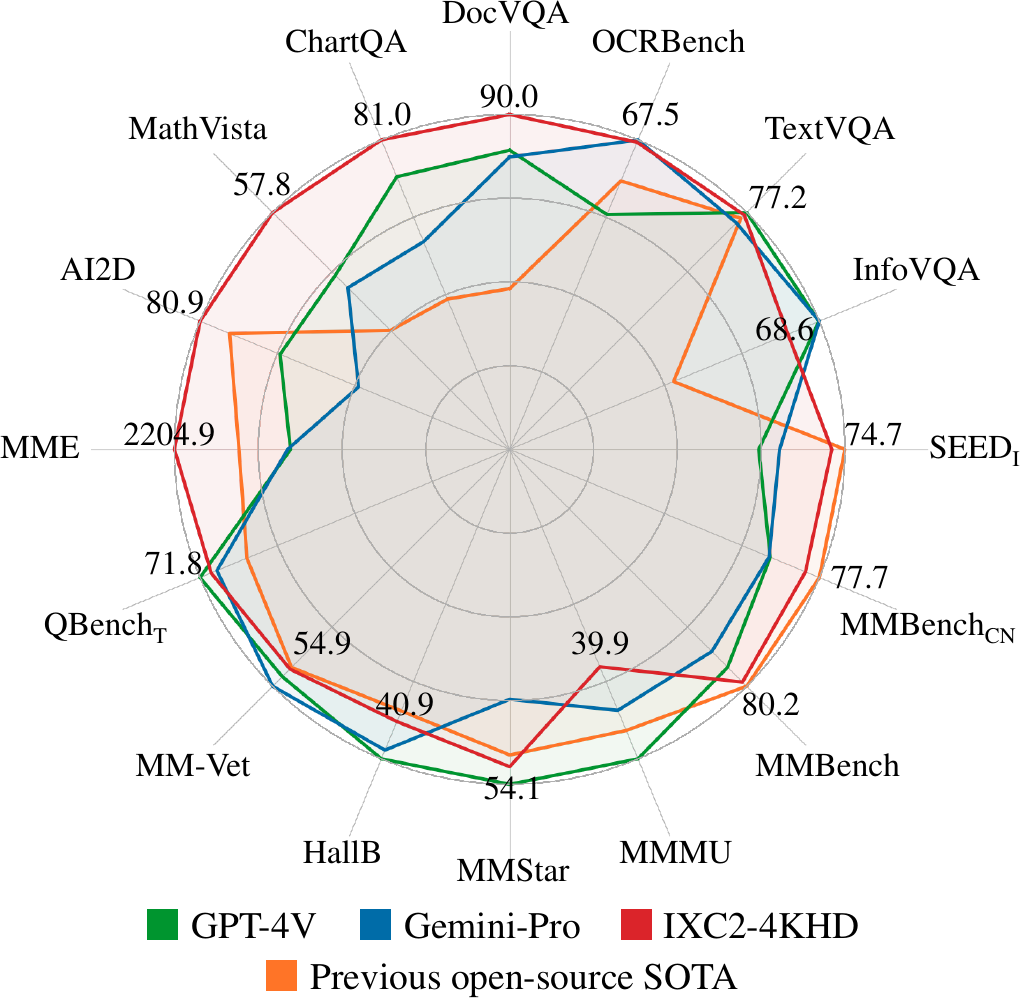}
    
    \setlength{\abovecaptionskip}{0mm} 
    \captionof{figure}{\small
        Overview of InternLM-XComposer2-4KHD performance on benchmarks with different resolutions.
        Our model based on InternLM2-7B~\cite{2023internlm} \textbf{matches or even surpasses GPT-4V~\cite{openai2023gpt4} and Gemini Pro~\cite{geminiteam2023gemini} in 10 of the 16 benchmarks}. 
	}
	\label{fig:teaser}
    %\vspace{-5pt}
\end{figure}

In recent years, the progress in Large Language Models (LLMs)~\cite{openai2020chatgpt,touvron2023llama,touvron2023llama2,jiang2023mistral, 2023internlm,cai2024internlm2,qwen7b,du2022glm,vicuna2023} has provoked the development of Large Vision-Language Models (LVLMs). These models have demonstrated proficiency in tasks such as image captioning~\cite{chen2015microsoft,chen2023sharegpt4v} and visual-question-answering (VQA)~\cite{MMBench,fu2023mme,seed_2023,yue2023mmmu}. Nevertheless, due to their limited resolution, they struggle with processing images containing fine details, such as charts~\cite{masry2022chartqa}, tables~\cite{textvqa}, documents~\cite{docvqa}, and infographics~\cite{infovqa}. This limitation constrains their practical applicability in real-world scenarios.

Recent advancements have aimed at enhancing the resolution of Large Vision-Language Models (LVLMs). Some approaches~\cite{lv2023kosmos25,cogagent,wei2023vary,li2024mini} involve adapting high-resolution vision encoders directly. However, the Vision Transformer (ViT) architecture falls short when dealing with images of varying resolutions and aspect ratios, thereby restricting its ability to handle diverse inputs effectively. Alternatively, some methods~\cite{li2023monkey,monkeytext,docowl,lin2023sphinx,llavauhd,llavanext,li2023otterhd} maintain the vision encoder's resolution, segmenting high-resolution images into multiple low-resolution patches. Yet, these methods are constrained by an inadequate resolution, typically around 1500 $\times$ 1500, which does not satisfy the demands of daily content, \eg, website screenshots~\cite{si2024design2code}, document pages~\cite{docvqa}, and blueprints~\cite{infovqa}. Furthermore, they are confined to either a few predefined high-resolution settings~\cite{cogagent, wei2023vary, li2024mini, li2023monkey, lin2023sphinx, llavanext,li2023otterhd,lv2023kosmos25,monkeytext} or a limited range of resolutions~\cite{docowl, llavauhd}, thereby restricting their utility across a variety of applications.

In this work, we introduce InternLM-XComposer2-4KHD, a pioneering model that for the first time expands the resolution capabilities of Large Vision-Language Models (LVLMs) to 4K HD and even higher, thereby setting a new standard in high-resolution vision-language understanding. Designed to handle a broad range of resolutions, InternLM-XComposer2-4KHD supports images with any aspect ratio from 336 pixels up to 4K HD, facilitating its deployment in real-world contexts.

InternLM-XComposer2-4KHD follows patch division~\cite{li2023monkey, li2023otterhd} paradigm and enhances it by incorporating an innovative extension: dynamic resolution with automatic patch configuration. To be specific, scaling the resolution of Large Vision-Language Models (LVLMs) to 4K HD and even higher standard is far beyond merely increasing the number of patches. It involves a nuanced approach to overcoming specific challenges: (1) \textbf{Dynamic Resolution and Automatic Patch Configuration}: Addressing the scarcity of high-resolution training data, our framework introduces a strategy that dynamically adjusts resolution alongside an automatic layout configuration. During training, it maintains the original aspect ratios of images while adaptively altering patch (336 $\times$ 336) layouts and counts. This results in a training resolution that exceeds the original image resolutions, reaching up to 4KHD,  addressing the shortfall of high-resolution data. (2) \textbf{Handling Variability in Patch Configurations}: Despite the apparent simplicity of dynamic resolution training, the variability in patch configurations can heavily confuse LVLMs. To mitigate this, we introduce a newline token after each row of patch tokens to clearly delineate patch layouts, reducing training ambiguity and significantly boosting performance. (3) \textbf{Inference Beyond 4K Resolution:} Our observations reveal that, even when trained on images up to 4K resolution, the model can achieve additional performance improvements during inference by processing images at higher resolutions. 

Furthermore, scaling the training resolution up to 4K standard results in a consistent improvement in performance, highlighting the potential for training even beyond 4K resolution. This underscores the capacity for further enhancing model capabilities and suggests a promising trajectory for advancing the frontiers of high-resolution image processing within the domain of large vision-language models.

We evaluate our InternLM-XComposer2-4KHD on 16 diverse benchmarks spanning various domains, including 5 challenging HD-OCR datasets (DocVQA\cite{docvqa}, ChartQA\cite{masry2022chartqa}, InfographicVQA\cite{infovqa}, TextVQA\cite{textvqa} and OCRBench\cite{ocrbench}). Compared to previous open-source LVLM models and closed-source APIs, our approach achieves SOTA results in 6 of 16 benchmarks, demonstrating competitive performance despite only 7B parameters. As shown in Figure~\ref{fig:teaser}, InternLM-XComposer2-4KHD even surpasses the performance of GPT4V~\cite{openai2023gpt4} and Gemini Pro~\cite{geminiteam2023gemini} across ten benchmarks.
Notably, our method exhibits excellent performance on 5 HD-OCR datasets, over existing open-source LVLMs by a substantial margin.

\begin{figure*}[h!]
    \centering
    \vspace{-30pt}
    \includegraphics[width=0.94\linewidth]{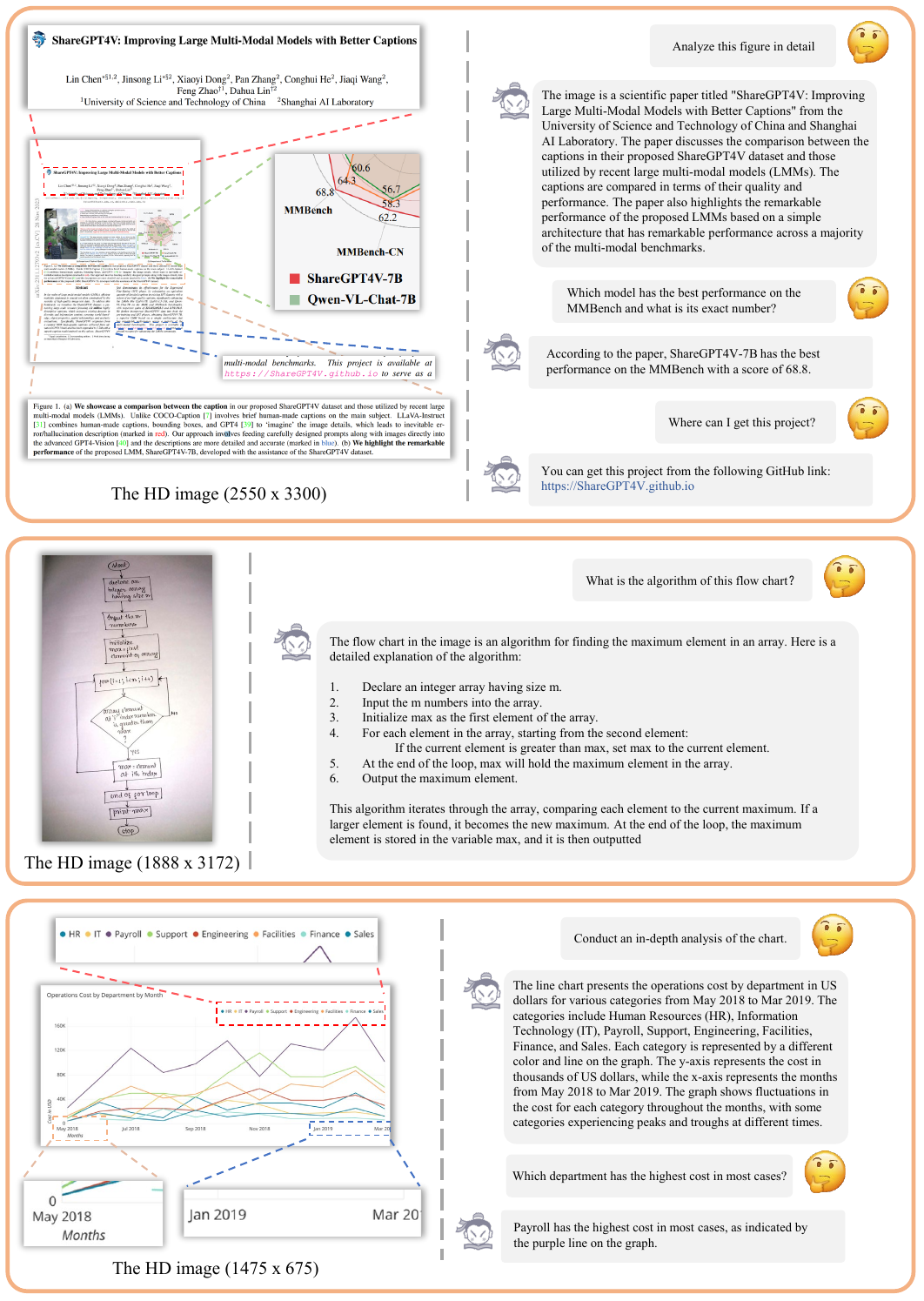}
    
    \setlength{\abovecaptionskip}{0mm} 
    \captionof{figure}{\small
        \textbf{Chat with InternLM-XComposer2-4KHD}. Some regions of the input HD images are zoomed in for better visualization.
	}
	\label{fig:teaser1}
    %\vspace{-5pt}
\end{figure*}

\begin{figure*}[h!]
    \centering
    \vspace{-40pt}
    \includegraphics[width=0.9\linewidth]{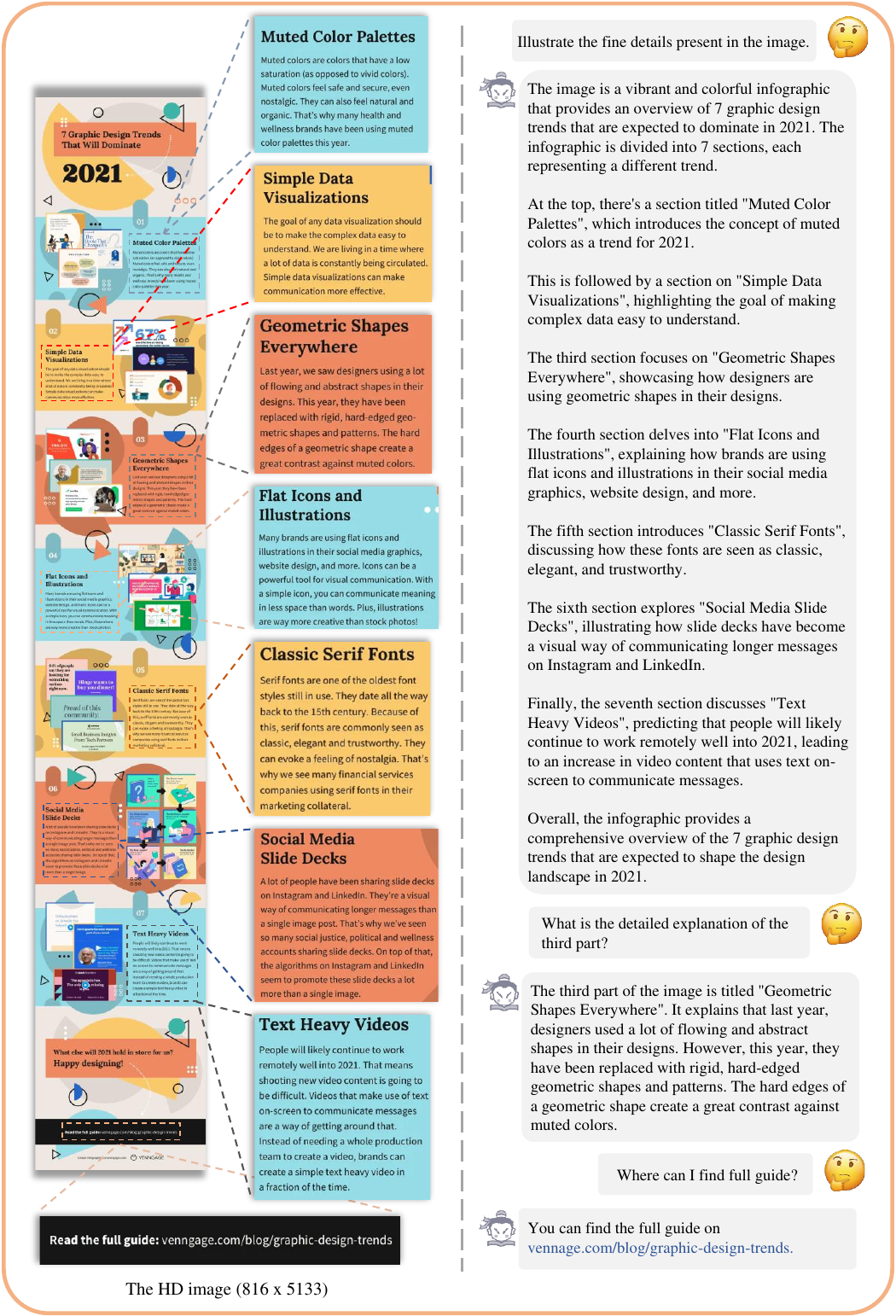}
    
    \setlength{\abovecaptionskip}{0mm} 
    \captionof{figure}{\small
        \textbf{Chat with InternLM-XComposer2-4KHD on ultra-high HD image with the 816 $\times$ 5133 resolution}. Some regions of the input HD images are zoomed in for better visualization.
	}
	\label{fig:teaser2}
    %\vspace{-5pt}
\end{figure*}

\iffalse

\twocolumn[{
\renewcommand\twocolumn[1][]{#1}
\maketitle
\begin{center}
    \centering
    \vspace{-20pt}
    \includegraphics[width=0.94\linewidth]{figures/teaser_cases.pdf}
    \setlength{\abovecaptionskip}{0mm}
    \captionof{figure}{\small
        Chat with InternLM-XComposer2-4KHD. Some regions of the input HD image are zoomed in for better visualization. 
	}
	\label{fig:teaser}
\end{center}
}]

\twocolumn[{
\renewcommand\twocolumn[1][]{#1}
\maketitle
\begin{center}
    \centering
    \vspace{-20pt}
    \includegraphics[width=0.91\linewidth]{figures/teaser_cases2.pdf}
    \setlength{\abovecaptionskip}{0mm}
    \captionof{figure}{\small
        Chat with InternLM-XComposer2-4KHD. Some regions of the input HD images are zoomed in for better visualization.
	}
	\label{fig:teaser}
\end{center}
}]
\fi

\section{Related Works}
\label{sec:related}

\noindent{\textbf{Large Vision-Language Models (LVLMs).}}
Large Language Models (LLMs)~\cite{brown2020language,ouyang2022training,openai2020chatgpt,chowdhery2022palm,kaplan2020scaling,touvron2023llama,touvron2023llama2,jiang2023mistral,2023internlm,zeng2023glm-130b,baichuan2023baichuan2,qwen7b,cai2024internlm2} have gained significant attention due to their impressive performance in various language-related tasks such as text generation and question answering. Following this enthusiasm, recent Large Vision-Language Models (LVLMs) have emerged\cite{openai2023gpt4,chen2023pali,chen2023palix,chen2023pali3,driess2023palme,fu2023gemini,zhu2023minigpt,dai2023instructblip,zhang2023internlm,fuyu-8b,li2023otter,peng2023kosmos,ye2023mplug,awadalla2023openflamingo}, combining LLMs with vision encoders~\cite{radford2021learning,zhang2024long,sun2023alpha} to leverage the complementary strengths of language and vision modalities. By fusing textual and visual representations, LVLMs can ground language in visual contexts, enabling a more comprehensive understanding and generation of multimodal content~\cite{chen2023sharegpt4v,chen2023internvl,lin2023sphinx,bai2023qwen,wang2023cogvlm,internlmxcomposer2,cao2024dualfocus,liu2024rar}.

\noindent \textbf{LVLMs for High-Resolution Understanding.}
Large Vision-Language Models (LVLMs) often employ CLIP-ViT as the visual encoder for vision-dependent tasks. However, the visual encoder's reliance on low resolutions, such as 224 $\times$ 224 or 336 $\times$ 336 pixels, limits its effectiveness for high-resolution tasks like OCR and document/chart perception. To enhance high-resolution understanding, recent works have primarily employed the following strategies:
(1) High-resolution (HR) visual encoders or dual encoders catering to HR and low-resolution (LR) inputs~\cite{lv2023kosmos25,wei2023vary,cogagent,li2024mini}. For instance, Vary~\cite{wei2023vary} introduces a new image encoder supporting HR inputs, which are then concatenated with LR embeddings from the original CLIP visual encoder. Similarly, CogAgent~\cite{cogagent} and Mini-Gemini~\cite{li2024mini} also separate HR and LR images using distinct vision encoders, subsequently merging their features using a cross-attention module. In contrast, our approach offers a more simplified solution and shows advantages for varying resolutions and aspect ratio inputs.
(2) Cropped image patches~\cite{li2023monkey,monkeytext,llavauhd,ureader,docowl,lin2023sphinx,li2023otterhd}. For example, Monkey~\cite{li2023monkey} employs sliding windows to segment images into patches, subsequently processing them with LoRA fine-tuning.
TextMonkey~\cite{monkeytext} further proposes shifted window attention and token resampler to consider the connections among different patches.
These approaches are confined to either a few predefined high-resolution settings~\cite{cogagent, wei2023vary, li2024mini, li2023monkey, lin2023sphinx, llavanext,li2023otterhd,lv2023kosmos25,monkeytext} or a limited range of resolutions~\cite{docowl, llavauhd}.
Conversely, our method devises a dynamic image partition strategy to support the scaling from 336 pixels to 4K resolution, and the maximum resolution is larger than previous approaches (\eg, 1.5k for Monkey~\cite{li2023monkey} and 2k for UReader~\cite{ureader}).

\noindent \textbf{LVLMs for Document Understanding.}
Document understanding involves analyzing and comprehending various digital documents, such as figures, tables, and academic papers. Many document understanding tasks require models to handle high-resolution inputs, complex layouts, various aspect ratios, and diverse document formats.
To enhance the capabilities of LVLMs for document understanding, several works have collected and constructed high-quality document instruction tuning data, including LLaVAR~\cite{zhang2023llavar}, mPLUG-DocOwl~\cite{ye2023mplug-doc} and TGDoc~\cite{wang2023towards}. DocPediaDocPedia~\cite{feng2023docpedia} processes document inputs in the frequency domain.
Some previous works have improved document understanding ability by designing special modules for high-resolution inputs, such as HR and LR encoders~\cite{cogagent,wei2023vary} or cropped image patches~\cite{ureader,monkeytext,llavauhd}.
Our InternLM-XComposer2-4KHD first scales to 4K resolution inputs and demonstrates strong document understanding ability on OCR-related benchmarks.
Also, our approach also achieves comparable results on other general LVLM benchmarks like perception and reasoning~\cite{lu2024mathvista,MMBench,seed_2023,mmstar}.
% method
\section{Method}

\begin{table*}[t]
\centering
\footnotesize
\setlength{\tabcolsep}{5mm}{
\begin{tabular}{ll}
\toprule
Task &  Dataset\\
\midrule
General Semantic Alignment  &  ShareGPT4V-PT~\cite{chen2023sharegpt4v}, COCO~\cite{chen2015microsoft},  Nocaps~\cite{agrawal2019nocaps}, TextCaps~\cite{sidorov2020textcaps},  LAION400M~\cite{schuhmann2021laion},  SBU~\cite{Ordonez_2011_im2text},   CC 3M~\cite{sharma2018conceptual} \\
World Knowledge Alignment & Concept Data~\cite{zhang2023internlm} \\
Vision Capability Enhancement & WanJuan~\cite{He2023WanJuanAC}, Flicker\cite{young2014flicker}, MMC-Inst\cite{liu2023mmc}, \textcolor{red}{ RCTW-17\cite{ocr_rctw}, CTW\cite{yuan2019ctw}, LSVT\cite{ocr_lsvt}, ReCTs\cite{ocr_rects}, ArT\cite{ocr_art}} \\
\bottomrule
\end{tabular}}
\vspace{-6pt}
\caption {\textbf{Datasets used for Pre-Training}. The data are collected from diverse sources for the three objectives. The newly added data is highlighted with \textcolor{red}{red}.}
\label{tab:pretrain_data}
\vspace{-12pt}
\end{table*}

\subsection{Model Architecture.}
The model architecture of InternLM-XComposer2-4KHD mainly follows the design of InternLM-XComposer2\cite{internlmxcomposer2} (XComposer2 in the following for simplicity.), including a light-weight Vision Encoder OpenAI ViT-Large/14, Large Language Model InternLM2-7B, and Partial LoRA for efficient alignment. We recommend the readers to the XComposer2 paper for more details.

\begin{figure}[t!]
    %\vspace{-7pt}
	\centering
	\includegraphics[width=1.\linewidth]{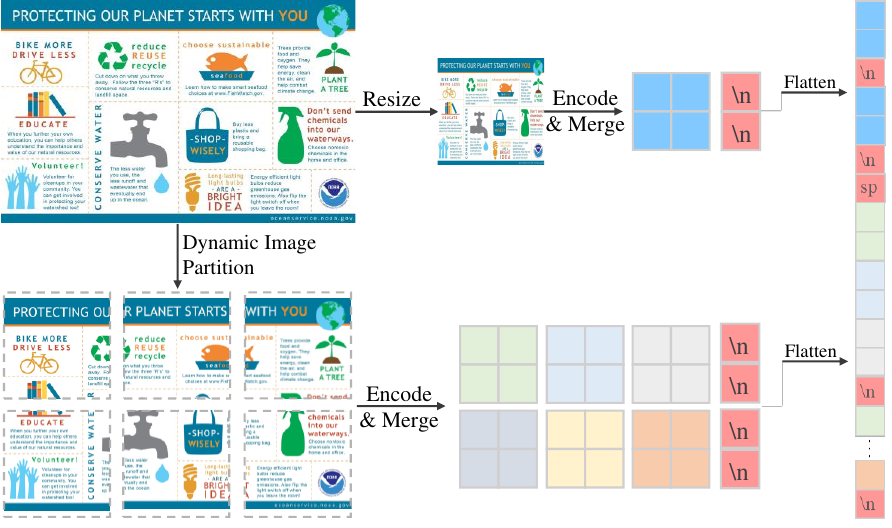}
    %\vspace{-2mm}
	\caption{\textbf{The illustration of processing high-resolution input.}  }
	\label{fig:framework}
 \vspace{-6pt}
\end{figure}

\subsection{High-Resolution Input.}

\noindent\textbf{Dynamic Image Partition.}
Utilizing a static input image size for processing high-resolution images, particularly those with varying aspect ratios, is neither efficient nor effective. To overcome this limitation, we introduce a dynamic image partitioning approach, as shown in Figure~\ref{fig:framework}. Our method strategically segments the image into smaller patches, while maintaining the integrity of the original image’s aspect ratio.

Given a maximum partition number $\mathcal{H}$, the image $x$ with size $[h,w]$ is resized and padded to the new image $\hat{x}$ with size $[p_h \times 336, p_w \times 336 ]$. This process is subject to the following constraints:
\begin{equation}     
      p_w \times p_h \leq \mathcal{H}; \;   p_h = \lceil p_w \times h / w \rceil 
\end{equation}
here $p_w$ and $p_h$ represent the number of patches in each row and column, respectively. We then split the $\hat{x}$ into $p_h \times p_w$ non-overlapped patches. Each patch is a small image with $336\times336$ size and we treat these patches as individual inputs for the ViT. 

In the following, we use `HD-$\mathcal{H}$' to represent our high-resolution setting with the constraint of $\mathcal{H}$ patches. For example, the 'HD-9' allows up to 9 patches, including a range of resolutions such as $1008\times1008$, $672\times1344$, $336\times3024$, \etc.

\noindent\textbf{Global-Local Format.}
For each input image, we present it to the model with two views. The first is the global view, where the image is resized to a fixed size (in our case, 336 × 336). This provides a macro understanding of the image. Empirically, we have found this to be crucial for the LVLM to correctly understand the image. The second view is the local view. We divide the image into patches using the previously mentioned Dynamic Image Partition strategy and extract features from each patch. Following feature extraction, the patches are reassembled into a large feature map. The feature map is then flattened to the final local features after a straightforward token merging process.

\noindent\textbf{Image 2D Structure Newline Indicator.}
Given that an image has a 2D structure and the image ratio is dynamic, the number of tokens for each row can vary across different images. This variation can potentially confuse the LVLM, making it difficult to determine which tokens belong to the same row of the image and which ones belong to the next row. This confusion may hinder the LVLM’s ability to understand the 2D structure of the image, which is crucial for comprehending structural image content such as documents, charts, and tables. To address this issue, we introduce a learnable newline (`$\backslash$n') token at the end of each row of the image features before the flattening.
Finally, we concatenate the global and local views, inserting a special `separate' token between them to distinguish the two views.

\subsection{Pre-Training}
During the pre-training phase, the LLM is frozen while both the vision encoder and Partial LoRA are fine-tuned to align the visual tokens with the LLM. The pre-training data mainly follow the design in XComposer2 which is curated with \textbf{three objectives} in mind:  1) general semantic alignment, 2) world knowledge alignment, 3) vision capability enhancement. 
In this paper, we focus on high-resolution and structural image understanding. Therefore, we have collected more related data to enhance this specific capability. 
As shown in Table.\ref{tab:pretrain_data}, we have utilized a diverse OCR dataset for this purpose.

In practice, we employ the OpenAI CLIP ViT-L-14-336 as the vision encoder. Different from XComposer2, We keep the ViT resolution as $336\times336$ and increase the input resolution with more patches. For the Dynamic Image Partition strategy, we use `HD-25' for the pertaining. For each image or patch, the image token number is decreased to $1/4$ with a simple \textbf{merge operation}. We concatenate the nearby 4 tokens into a new token through the channel dimension, then align it with the LLM by an MLP. The `separate' and `$\backslash$n' token are randomly initialized.
For the Partial LoRA, we set a rank of $256$ for all the linear layers in the LLM decoder block. 
Our training process involves a batch size of 4096 and spans across 2 epochs. The learning rate linearly increases to $2 \times 10^{-4}$ within the first $1\%$ of the training steps. Following this, it decreases to $0$ according to a cosine decay strategy. 
To preserve the pre-existing knowledge of the vision encoder, we apply a layer-wise learning rate (LLDR) decay strategy, and the decay factor is set to $0.90$.

\begin{table}[t]
\centering
\footnotesize
\setlength{\tabcolsep}{1mm}{
\begin{tabular}{ll}
\toprule
Task &  Dataset\\
\midrule 
Caption  &  ShareGPT4V~\cite{chen2023sharegpt4v}, COCO~\cite{chen2015microsoft},Nocaps~\cite{agrawal2019nocaps} \\\midrule
General QA  & VQAv2~\cite{VQAv2}, GQA~\cite{hudson2018gqa}, OK-VQA~\cite{marino2019ok} \\
            & VD \cite{visdial}, RD\cite{chen2023shikra}, VSR\cite{Liu2022VisualSR}, \\ \midrule
Science QA  & AI2D~\cite{kembhavi2016diagram}, SQA~\cite{lu2022learn}, TQA\cite{tqa}, IconQA\cite{lu2021iconqa}\\\midrule
Chart QA    & DVQA~\cite{kafle2018dvqa}, ChartQA, \textcolor{red}{ChartQA-AUG~\cite{masry2022chartqa}} \\\midrule
Math QA     & MathQA~\cite{yu2023metamath}, Geometry3K\cite{Geometry3K}, TabMWP\cite{tabmwp}, \\
&    CLEVR-MATH\cite{clevr_math}/Super\cite{clevr_super} \\\midrule
World Knowledge QA & A-OKVQA~\cite{schwenk2022okvqa},KVQA~\cite{shah2019kvqa}, ViQuAE\cite{lerner2022viquae} \\\midrule
\textcolor{red}{OCR QA} & \textcolor{red}{TextVQA\cite{textvqa}, OCR-VQA\cite{ocr_vqa}, ST-VQA\cite{stvqa}} \\\midrule
\textcolor{red}{HD-OCR QA} & \textcolor{red}{InfoVQA\cite{infovqa},  DocVQA\cite{docvqa}}\\\midrule
Conversation & LLaVA-150k~\cite{liu2023visual}, LVIS-Instruct4V~\cite{wang2023see} \\
& ShareGPT-en\&zh ~\cite{vicuna2023}, InternLM-Chat\cite{2023internlm} \\ 
\bottomrule
\end{tabular}}
\vspace{-6pt}
\caption {\textbf{Datasets used for Supervised Fine-Tuning}. We collect data from diverse sources to empower the model with different capabilities. The newly added data is highlighted with \textcolor{red}{red}.}
\label{tab:sft data}
\vspace{-12pt}
\end{table}

\begin{table*}[t!]
\footnotesize
\centering
\setlength{\tabcolsep}{1.3mm}{
\begin{tabular}{l|cccccccccccccccc}
\toprule

\multirow{ 2}{*}{Method} & Doc & Chart & Info & Text & OCR & MM & Math & \multirow{ 2}{*}{AI2D} & \multirow{ 2}{*}{MMMU} & \multirow{ 2}{*}{MME} & MMB & MMB & SEED & QBench & MM- & Hall    \\ 
 &   VQA & QA & VQA & VQA & Bench & Star & Vista & ~ & ~ & ~ & EN & CN & Image & Test & Vet & Bench \\ 
\midrule 
Open-Source  &  \cite{docowl} & \cite{docowl} & \cite{docowl} & \cite{cogagent} & \cite{cogagent} & \cite{llavanext} & \cite{llavanext} & \cite{llavanext} & \cite{chen2023internvl} & \cite{ai2024yi} & \cite{llavanext} & \cite{llavanext} & \cite{llavanext} & \cite{zhang2023internlm} & \cite{wang2023cogvlm} & \cite{li2023monkey}   \\ 
Previous SOTA  & 8B & 8B & 8B & 18B & 18B & 35B & 35B & 35B & 40B & 34B & 35B & 35B & 35B & 8B & 17B & 10B  \\ 
  & 82.2 & 70.2 & 44.5 & 76.1 & 59.0 & 52.1 & 39.0 & {78.9} & \underline{51.6} & {2050.2} & \textbf{81.1} & \underline{79.0} & \textbf{75.7} & 64.4 & 54.5 & 39.3  \\  
 \midrule \multicolumn{11}{l}{\textit{Closed-source API}} \\
        GPT-4V & \underline{88.4} & \underline{78.5} & \underline{75.1} & \textbf{78.0} & 51.6 & \textbf{57.1} & \underline{47.8} & 75.5 & \textbf{56.8} & 1,926.5 & 77.0 & 74.4 & 69.1 & \textbf{74.1} & \underline{56.8} & \textbf{46.5}   \\ 
        Gemini-Pro &88.1 & 74.1 & \textbf{75.2} & 74.6 & \textbf{68.0} & 42.6 & 45.8 & 70.2 & {47.9} & 1,933.3 & 73.6 & 74.3 & 70.7 & 70.6 & \textbf{59.2} & \underline{45.2}  \\ 
\midrule
\rowcolor[HTML]{F2F3F5} 
        IXC2-VL & 57.7 & 72.6 & 34.4 & 70.1 & 53.2 & \underline{55.4} & 57.6 & \textbf{81.2} & 41.4 & \textbf{2,220.4} & \underline{80.7} & \textbf{79.4} & 74.9 & \underline{72.5} & 46.7 & 41.0 \\  
\rowcolor[HTML]{F2F3F5} 
        IXC2-4KHD & \textbf{90.0} & \textbf{81.0} & 68.6 & \underline{77.2 }& 67.5 & {54.1} & \textbf{57.8} & \underline{80.9} & 39.7 & \underline{2,204.9} & {80.2} & {77.7} & \underline{74.7} & {71.8} & 54.9 & 40.9 \\

 \bottomrule
\end{tabular} }
\vspace{-2mm}
\caption{\textbf{Comparison with closed-source APIs and previous open-source SOTAs.} Our InternLM-XComposer2-4KHD gets SOTA results in 6 of the 16 benchmarks with only 7B parameters, showing competitive results with current closed-source APIs. The best results are \textbf{bold} and the second-best results are \underline{underlined}.}
\vspace{-3mm}
\label{tab:sota_comp}
\end{table*}

\subsection{4KHD Supervised Fine-tuning}
After the pre-training, we empower the model to understand high-resolution images and solve diverse challenges. 
Different from previous perception tasks (\eg, VQAv2, GQA) which typically answer questions based on the noticeable object in the image. OCR-related tasks depend on a detailed understanding of text within a high-resolution image. For instance, in InfoVQA, the length of the longer side of 50\% of the images exceeds 2000 pixels. Low-resolution inputs can distort the dense text information, causing the model to fail in its understanding. However, we have observed a resolution saturation problem with the aforementioned perception tasks, where the influence of resolution becomes negligible.

\begin{table*}[!ht] 
    \small
    \centering
    \setlength{\tabcolsep}{1.2mm}{
    \begin{tabular}{l|l|cccccccccc}
    \toprule
        Method & LLM & MMStar & MathVista & AI2D & MME$^P$ & MME$^C$ & MMB & MMB$^{CN}$ & SEED$^I$ & QBench$^T$ & MM-Vet  \\ 
    \midrule
        Qwen-VL-Chat & Qwen-7B & 37.5 & 33.8 & 63.0 & 1,487.5 & 360.7 & 60.6 & 56.7 & 58.2 & 61.7 & 47.3  \\ 
        
        ShareGPT4V  & Vicuna-7B & 33.0 & 25.8 & 58.0 & 1,567.4 & 376.4 & 68.8 & 62.2 & 69.7 & - & 37.6  \\ 
        Monkey & Qwen-7B & 38.3 & 34.8 & 62.5 & 1,522.4 & {401.4} & 72.4 & 67.5 & 68.9 & - & 33.0  \\ 
        CogVLM-17B & Vicuna-7B & 36.5 & 34.7 & 63.3 & - & - & 65.8 & 55.9 & 68.8 & - & \underline{54.5}  \\ 
        LLaVA-XTuner & InernLM2-20B & - & 24.6 & 65.4 & - & - & 75.1 & 73.7 & 70.2 & - & 37.2  \\ 
        LLaVA-1.5 & Vicuna-13B & 32.8 & 26.1 & 61.1 & 1,531.3 & 295.4 & 67.7 & 63.6 & 68.2 & 61.4 & 35.4  \\ 
        LLaVA-Next & Vicuna-13B &  38.3 & 32.4 & 72.2 & 1,445.0 & 296.0 & 70.0 & 68.5 & 71.4 & - & 44.9  \\ 
        InternLM-XC & InernLM-7B & - & 29.5 & 56.9 & 1,528.4 & 391.1 & 74.4 & 72.4 & 66.1 & {64.4 }& 35.2  \\ 
        \midrule
        \rowcolor[HTML]{F2F3F5} IXC2-VL & InernLM2-7B & \textbf{55.4} & \underline{57.6} & \textbf{81.2} & \textbf{1,712.0} & \underline{530.7} & \textbf{80.7} & \textbf{79.4} & \textbf{74.9} & \textbf{72.5} & 46.7  \\
        \rowcolor[HTML]{F2F3F5} IXC2-4KHD & InernLM2-7B & \underline{54.1} & \textbf{57.8} & \underline{80.9} & \underline{1,655.9} & \textbf{548.9} & \underline{80.2} & \underline{77.7} & \underline{74.7} & \underline{71.8} & \textbf{54.9} \\ 
    \bottomrule
    \end{tabular}}
    \vspace{-2mm}
\caption{\textbf{Comparison with open-source SOTA methods.} IXC2-4KHD outperforms competitors in most benchmarks. The best results are \textbf{bold} and the second-best results are \underline{underlined}.}
\label{tab:entire_comp}
\vspace{-6pt}
\end{table*}

To address this, we introduce a mixed-resolution training strategy for more efficient training. For tasks requiring high resolution, we employ the `HD-55' setting during training. This allows for the input of 4K ($3840\times1600$) images without necessitating additional image compression. These tasks are referred to as the HD-OCR QA tasks in Table~\ref{tab:sft data}. 
For other tasks, we implement a dynamic-resolution strategy. Images are resized to fall within a range between their original size and the size specified by the `HD25' setting. This dynamic approach enhances the robustness of the LVLM against differences in input resolution, thereby enabling the LVLM to utilize a larger resolution during inference. For instance, we have observed that using the `HD30’ setting yields better results on most OCR-related tasks when the LVLM is trained under the `HD25’ setting.

In practice, we jointly train all the components with a batch size of 2048 over 3500 steps. % Comment: Joint training details
Data from multiple sources are sampled in a weighted manner, with the weights based on the number of data from each source. 
As the `HD-55' setting has double image tokens than the `HD-25', we adjust the data loader to enable different batch sizes for them and adjust their weight accordingly.
The maximum learning rate is set to $5 \times 10^{-5}$, and each component has its own unique learning strategy. % Comment: Learning rate and strategy details
For the vision encoder, we set the LLDR to $0.9$, which aligns with the pretraining strategy. % Comment: Vision encoder learning strategy
For the LLM, we employ a fixed learning rate scale factor of $0.2$. This slows down the update of the LLM, achieving a balance between preserving its original capabilities and aligning it with vision knowledge. % Comment: LLM learning strategy

\section{Experiments}
In this section, we validate the benchmark performance of our InternLM-XComposer2-4KHD (IXC2-4KHD in the following for simplicity) after supervised fine-tuning.

\begin{figure*}[t]
    %\vspace{-7pt}
	\centering
	\includegraphics[width=1.\linewidth]{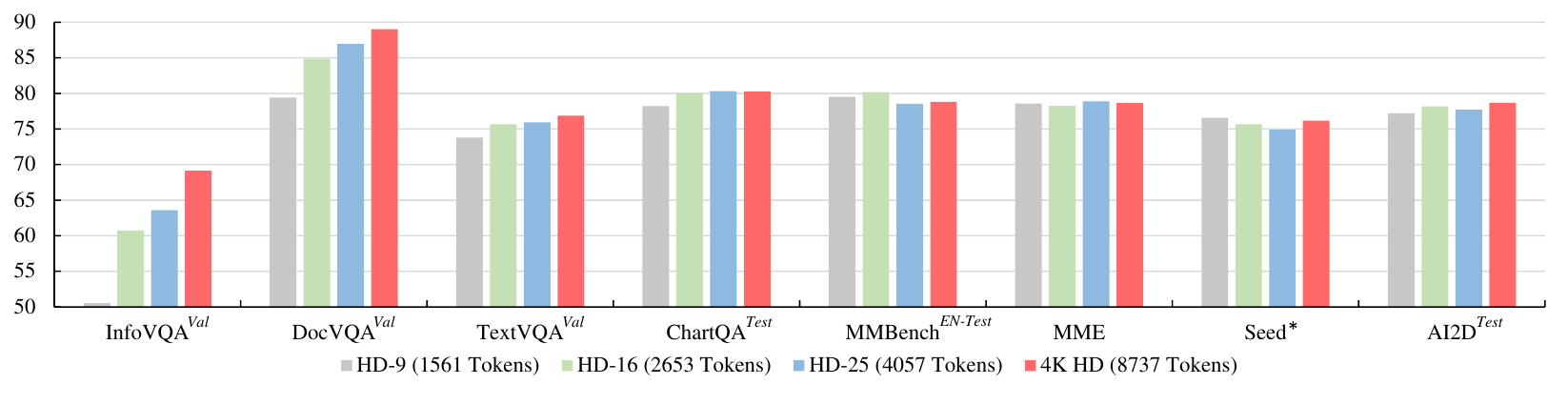} 
        \vspace{-10mm}
	\caption{\textbf{Influence of Training Resolution.} High-resolution training is critical for HD-OCR tasks, while its gain on other tasks is minor. }
	\label{fig:reso}
    \vspace{-6pt}
\end{figure*}

\begin{table*}[!ht]
    \small
    \centering
    \setlength{\tabcolsep}{1.5mm}{
    \begin{tabular}{l|l|l|lllll}
    \toprule
        Model & Model Size &Max Resolution & DocVQA$^{Test}$ & ChartQA$^{Test}$ & InfoVQA$^{Test}$ & TextVQA$^{Val}$ & OCRBench  \\ \midrule
        
        TextMonkey\cite{monkeytext} & 9B & 896x896 & 73.0 & 66.9 & 28.6 & 65.6 & 55.8  \\ 
        LLaVA-UHD \cite{llavauhd} & 13B & 1008x672 & --- & --- & --- & 67.7 & ---  \\ 
        CogAgent \cite{cogagent} & 17B & 1024x1024 & 81.6 & 68.4 & 44.5 & 76.1 & 59.0  \\ 
        UReader \cite{ureader}  & 7B  &2240x2240 & 65.4 & 59.3 & 42.2 & 57.6 & --- \\ 
        DocOwl 1.5 \cite{docowl} & 8B &  1344x1344 & 82.2 & 70.2 & 50.7 & 68.6 & ---  \\ 
        
        \midrule \rowcolor[HTML]{F2F3F5} 
        IXC2-4KHD & 8B &3840x1600 & \textbf{90.0} (+7.8) & \textbf{81.0} (+10.8) &\textbf{68.6 }(+17.9) &\textbf{77.2} (+1.2) & \textbf{67.5 }(+8.5)\\ 
        \bottomrule
    \end{tabular}}
    \vspace{-6pt}
    \caption{\textbf{High-resolution Evaluation.} IntenrLM-XComposer2-4KHD has the largest input resolution and outperforms open-source LVLMs which are specifically tuned for document understanding. }
\vspace{-4mm}
\label{tab:high-reso}
\end{table*}

\subsection{LVLM Benchmark results.}
In Table \ref{tab:sota_comp} and Table \ref{tab:entire_comp}, we compare our IXC2-4KHD on a list of benchmarks with both SOTA open-source LVLMs and closed-source APIs. Here we report results in DocVQA\cite{docvqa}, ChartQA\cite{masry2022chartqa}, InfographicVQA\cite{infovqa}, TextVQA\cite{textvqa}, OCRBench\cite{ocrbench}, MMStar\cite{mmstar}, 
MathVista\cite{lu2024mathvista}, MMMU\cite{yue2023mmmu}, AI2D\cite{kembhavi2016diagram}, MME \cite{fu2023mme}, MMBench (MMB) \cite{MMBench}, MMBench-Chinese (MMB$^{CN}$) \cite{MMBench}, SEED-Bench Image Part (SEED$^{I}$)\cite{li2023seedbench}, QBench-Testset (QBench$^{T}$)\cite{wu2023q}, MM-Vet \cite{yu2023mmvet}, HallusionBench (HallB)\cite{guan2023hallusionbench}. The evaluation is mainly conducted on the OpenCompass VLMEvalKit\cite{2023opencompass} for the unified reproduction of the results.

\noindent\textbf{Comparison with Closed-Source APIs.}
\noindent As demonstrated in Table \ref{tab:sota_comp},  IXC2-4KHD exhibits competitive performance across a variety of benchmarks, rivaling that of Closed-Source APIs. Owing to its high-resolution input,  IXC2-4KHD achieves a score of $90.0\%$ on DocVQA and $81.0\%$ on ChartQA, thereby surpassing GPT-4V and Gemini-Pro with a non-trivial margin. In the challenging InfographicVQA task, our model is the first open-source model that is close to the performance of Closed-Source APIs, exceeding the performance of previous open-source models by nearly $20\%$. In addition to OCR-related tasks, IXC2-4KHD is a general-purpose Large Vision-Language Modal that excels in semantic-level tasks, demonstrating competitive results.

\noindent\textbf{Comparison with Open-Source Models.}
\noindent We also conduct a comprehensive comparison with open-source LVLMs under a similar model scale. As shown in Table \ref{tab:entire_comp}, our model significantly outperforms existing open-source models, achieving competitive results across all benchmarks. Notably, the InternLM-XComposer2 series is the only method that achieves a higher than $50\%$ score on the challenging MMStar benchmark.

\noindent\textbf{High-resolution Understanding Evaluation.}
\noindent Then we compare IXC2-4KHD with models that are specifically designed for high-resolution understanding tasks. We report the results of 5 high-resolution benchmarks in Table \ref{tab:high-reso}, as a general LVLM, IXC2-4KHD shows superb performance on these tasks and outperforms competitors with a large margin. For example, IXC2-4KHD gets $68.6\%$ on InfographicVQA, surpassing recent DocOwl 1.5 with $+17.9\%$. For the OCRBench, IXC2-4KHD gets $67.5\%$, outperforms CogAgent with $+8.5\%$.

\subsection{Dive into Resolution}
\noindent\textbf{High-Resolution Training is Critical for HD-OCR tasks.}
We study four resolution settings: HD-9 (1561 image tokens at most, we simply the statement if the following), HD-16 (2653 tokens), HD-25 (4057 tokens), and 4KHD (8737 tokens). Here we report the validation set of InfoVQA, DocVQA, and TextVQA, test set of ChartQA and AI2D, MMBench EN-Test, and a 2k subset of SEEDBench (we denote it as SEED$^*$). In the following experiments, we report results on the above benchmarks by default.

As illustrated in Fig.\ref{fig:reso}, we note a significant improvement in the HD-OCR tasks as the resolution increases. For instance, the model achieves only a $50.5\%$ score on the InfographicVQA with the HD-9 setting. However, when we switch to the HD-16 setting, we observe a performance gain of $+10.2\%$. The performance continues to improve as the resolution increases, with saturation not observed even for the 4KHD setting. Due to computational constraints, we defer the exploration of the upper bound of improvement to future work.
In terms of other OCR-related tasks, the performance gain attributable to increased resolution is relatively minor.
For the perception-related benchmarks, performance is saturated on the resolution that only has negligible difference between the four settings.

\begin{table}[t]
    \centering
    \footnotesize
    \setlength{\tabcolsep}{1.3mm}{
    \begin{tabular}{ll|ccccccc}
    \toprule
        Train & Eval &  Doc & Info &  Text & Chart & MMB & MME & SEED$^*$  \\ 
    \midrule
    \multirow{ 2}{*}{HD9} & HD9 & 79.4 & 50.5 & 73.8 & 78.2 & 79.5 & 2,201 & 76.6  \\ 
        ~ & HD16 & 83.0 & 58.6 & 74.3 & 75.8 & 79.3 & 2,198 & 76.7  \\ 
    \midrule
    \multirow{ 2}{*}{HD16} & HD16 & 84.9 & 60.8 & 75.7 & 80.1 & 80.2 & 2,129 & 75.7  \\ 
        ~ & HD25 & 85.9 & 62.1 & 75.8 & 79.1 & 80.1 & 2,100 & 75.4 \\ 
    \midrule
        \multirow{ 2}{*}{HD25} & HD25 & 87.0 & 63.6 & 76.0 & 80.3 & 78.5 & 2,209 & 74.9  \\ 
        ~ & HD30 & 87.4 & 64.6 & 76.2 & 79.4 & 78.9 & 2,173 & 74.3  \\ 
    \bottomrule
    \end{tabular}}\vspace{-2mm}
    \caption{\textbf{Influence of Inference Resolution.} The model achieves better performance on text-related tasks when the inference resolution is higher than its training resolution.} 
    \vspace{-4mm}
\label{tab:eval_resolution}
\end{table}

\noindent\textbf{Higher Inference Resolution Leads to better results on Text-related Tasks.}
An intriguing observation from our experiments is that our model, when inferring with a slightly higher resolution, tends to yield improved results on text-related tasks. We present the results of HD-9, HD-16, and HD-25 in Table \ref{tab:eval_resolution}. For instance, IXC2-HD9 achieves a $50.5\%$ score on InfographicVQA. When we infer with HD16, we see a performance gain of $+8.1\%$, without additional training. Similar improvements are also observed with IXC2-HD16 and IXC2-HD25. We posit that the dynamic image token length used in training enhances the robustness of the LVLM, leading to better results when the text in the image is more `clear' in the higher resolution input.
Conversely, the results on ChartQA consistently degrade under this setting. This could be due to the model becoming confused about the chart structure when the resolution is altered. 
Additionally, similar to the observation from Figure \ref{fig:reso}, the impact of resolution on perception-related benchmarks appears to be quite minor.

\noindent\textbf{Visualization Results.} We provide the visualization results on ultra-high HD images in Figure \ref{fig:teaser1} and Figure \ref{fig:teaser2}. Please refer to the appendix for more results.

\subsection{High-Resolution Strategy Ablation}
\noindent\textbf{The Role of Global-View.}
We first examine the impact of the global view in our Global-Local Format. As indicated in Table \ref{tab:global_view}, we find that the global view is essential for the LVLM to accurately comprehend the input image. When it is removed, the model performs worse across all benchmarks. For instance, the model experiences a $-4.4\%$ drop in performance on the MMBench EN-Test without the global view. We contend that the global view offers a general macro understanding of the image, which the model struggled to derive from the large number of tokens in the local view.

\begin{table}[t]
    \centering
    \footnotesize
    \setlength{\tabcolsep}{1.mm}{
    \begin{tabular}{lccccccc}
    \toprule
        Model & Doc & Info & Text& Chart  & MMB & MME & SEED$^*$  \\ 
    \midrule
        HD9 & 79.4 & 50.5 & 73.8& 78.2  & 79.5 & 2201 & 76.6  \\ 
        + w/o global-view & 78.1 & 47.9 & 71.2 & 77.9 & 75.1 & 2019 & 76.2 \\ 
    \bottomrule
    \end{tabular}}\vspace{-2mm}
    \caption{\textbf{Influence of Global-View in the Input.} Global-view is critical for most benchmarks. } 
\label{tab:global_view}
\vspace{-6pt}
\end{table}

\noindent\textbf{The Role of the Newline Token.}
We incorporate a special newline token at the end of each row of the image features before the flattening operation. This token serves as an indicator of the image’s 2D structure. We examine its impact on both the HD-9 and 4KHD strategies in Table \ref{tab:gang_n}.
When a fixed high-resolution strategy HD-9 is employed, we observe that the benefit derived from the newline token is minor. This could be attributed to the LVLM’s ability to handle limited differences in image ratios after training. However, when we implement a more challenging 4KHD (HD-25 + HD-55) strategy, which exhibits significant diversity in both image ratio and token number, the LVLM demonstrates a notable decline in performance on OCR-related tasks without the newline indicator.
This finding supports our hypothesis that the LVLM struggles to comprehend the shape of the image when the image tokens are directly flattened into a 1D sequence. The newline token can assist the model in better understanding the structure of the image.

\begin{table}[t]
    \centering
    \footnotesize
    \setlength{\tabcolsep}{1.5mm}{
    \begin{tabular}{lcccccccc}
    \toprule
        Model & `$\backslash$n' & Doc & Info  & Text & Chart& MMB & MME & SEED$^*$  \\ 
    \midrule
        HD9 & $\times$ & 79.5 & 50.3  & 74.0& 78.2 & 79.1 & 2206 & 75.9  \\ 
        HD9 & \checkmark & 79.4 & 50.5  & 73.8& 78.2 & 79.5 & 2201 & \textbf{76.6}  \\ 
    \midrule
        4KHD & $\times$ & 88.1 & 67.4 & 75.9& 80.4  & 79.9 & \textbf{2232} & 76.4  \\ 
        4KHD & \checkmark & \textbf{89.0} & \textbf{69.3}& \textbf{77.2} & \textbf{81.0}  & \textbf{80.2} & 2205 & 76.2 \\ 
    \bottomrule
    \end{tabular}}\vspace{-2mm}
    \caption{\textbf{Influence of Indicator `$\backslash$n'  in the Image Features.} `$\backslash$n' helps LVLM understand structural images when the input resolution is dynamic and large. }
    \vspace{-6pt}
\label{tab:gang_n}
\end{table}

\noindent\textbf{Influence of Token Merging Strategy.}
In practice, we employ a simple merging strategy that concatenates four adjacent tokens along the channel dimension. We have found this approach to be effective in reducing the number of image tokens efficiently. Here we study the influence of different token-merging strategies under the 4KHD setting. In Table \ref{tab:merge}, we study two additional strategies: Re-Sampler\cite{bai2023qwen} and C-Abstractor\cite{cha2023honeybee}, with their default setting and the same compressing rate $0.25$, \ie, reducing an image with 576 tokens to 144 tokens. 
Results show that both concatenation and C-Abstractor work well and get similar results on most benchmarks, this observation is also consistent with the study in MM-1\cite{mckinzie2024mm1} that the influence of the connector is minor.
However, the Re-Sampler performs worse than the other methods with a noticeable margin. We argue this is caused by the learnable queries used for gathering information requiring a great number of data for training, our pre-training data is somewhat lightweight for it to converge fully.

\begin{table}[t]
    \centering
    \footnotesize
    \setlength{\tabcolsep}{1.5mm}{
    \begin{tabular}{lccccccc}
    \toprule
          Strategy & Doc & Info  & Text & Chart& MMB & MME & SEED$^*$  \\
    \midrule
        Re-Sampler   & 86.2 & 67.1 & 75.3 & 78.8 & 79.6 & 2124 & 74.2 \\
        C-Abstractor & 88.6 & 69.5 & 77.1 & 80.6 & 80.4 & 2236 & 76.7 \\
        Concat       & 89.0 & 69.3 & 77.2 & 81.0 & 80.2 & 2205 & 76.2 \\ 
    \bottomrule
    \end{tabular}}\vspace{-2mm}
    \caption{\textbf{Ablation on Token Merging Strategy.} Both the simple concatenation operation and the C-Abstractor works well.}
    \vspace{-8pt}
\label{tab:merge}
\end{table}

\section{Conclusion}
In this paper, we propose the InternLM-Xcomposer2-4KHD that exceeds the performance of previous open-source models on OCR-related tasks and also achieves competitive results on general-purpose LVLM benchmarks. Thanks to our dynamic resolution and automatic patch configuration, our model supports a maximum training resolution of up to 4K HD. We also integrate a global view patch to support the macro understanding and a learnable newline token to handle the various input image resolutions. Our model's performance continues to improve as the training resolution increases for HD-OCR tasks. Notably, we do not observe any performance saturation even for the 4KHD setting, and we have not explored the upper bound due to the computational burden increasing with higher-resolution inputs. In future work, we plan to explore efficient solutions for accurate LVLM training and inference, enabling our model to handle even higher resolutions while maintaining computational efficiency.

 \newpage
 \clearpage
%%%%%%%%% REFERENCES
{\small
\bibliographystyle{ieee_fullname}
\bibliography{egbib}
}

\appendix
\newpage

\twocolumn[{
\renewcommand\twocolumn[1][]{#1}
\maketitle
\begin{center}
    \centering
    \vspace{-20pt}
    \includegraphics[width=0.91\linewidth]{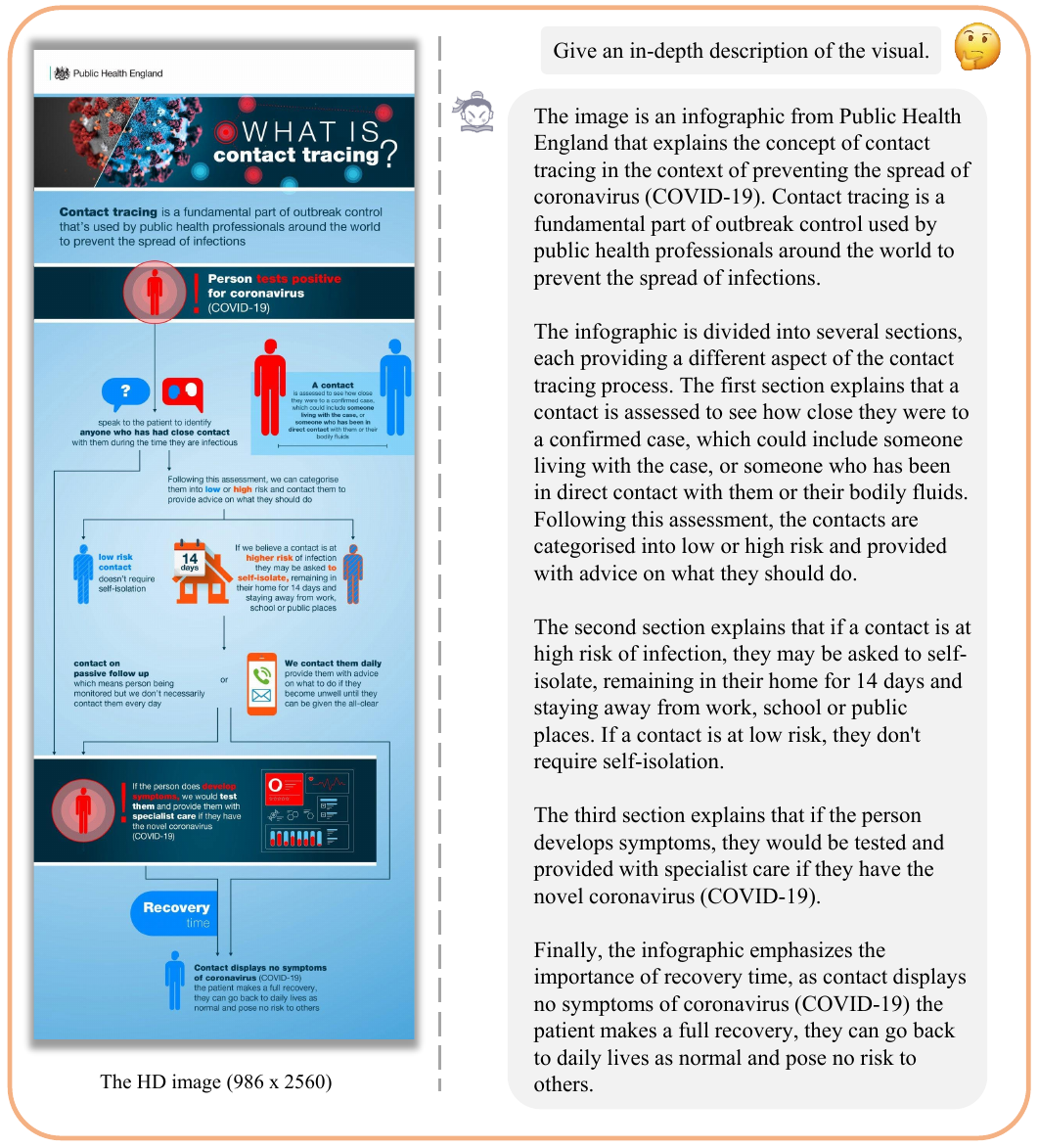}
    \setlength{\abovecaptionskip}{0mm}
    \captionof{figure}{\small
        \textbf{Chat with InternLM-XComposer2-4KHD}. Please refer to the \href{https://scontent-xsp1-2.xx.fbcdn.net/v/t1.6435-9/90180908_2952545558121906_3228929853055041536_n.jpg?stp=dst-jpg_p526x296&_nc_cat=104&ccb=1-7&_nc_sid=5f2048&_nc_ohc=cBeZG37pVNQAX_vRTdv&_nc_ht=scontent-xsp1-2.xx&oh=00_AfBPHBcZ6Qk6l_VD42rDoHZMS9IRn_YJ7EyS-MQEGX1Q_A&oe=663480F2}{link} for better visualization.
	}
	\label{fig:teaser_7}
\end{center}
}]

\twocolumn[{
\renewcommand\twocolumn[1][]{#1}
\maketitle
\begin{center}
    \centering
    \vspace{-20pt}
    \includegraphics[width=0.91\linewidth]{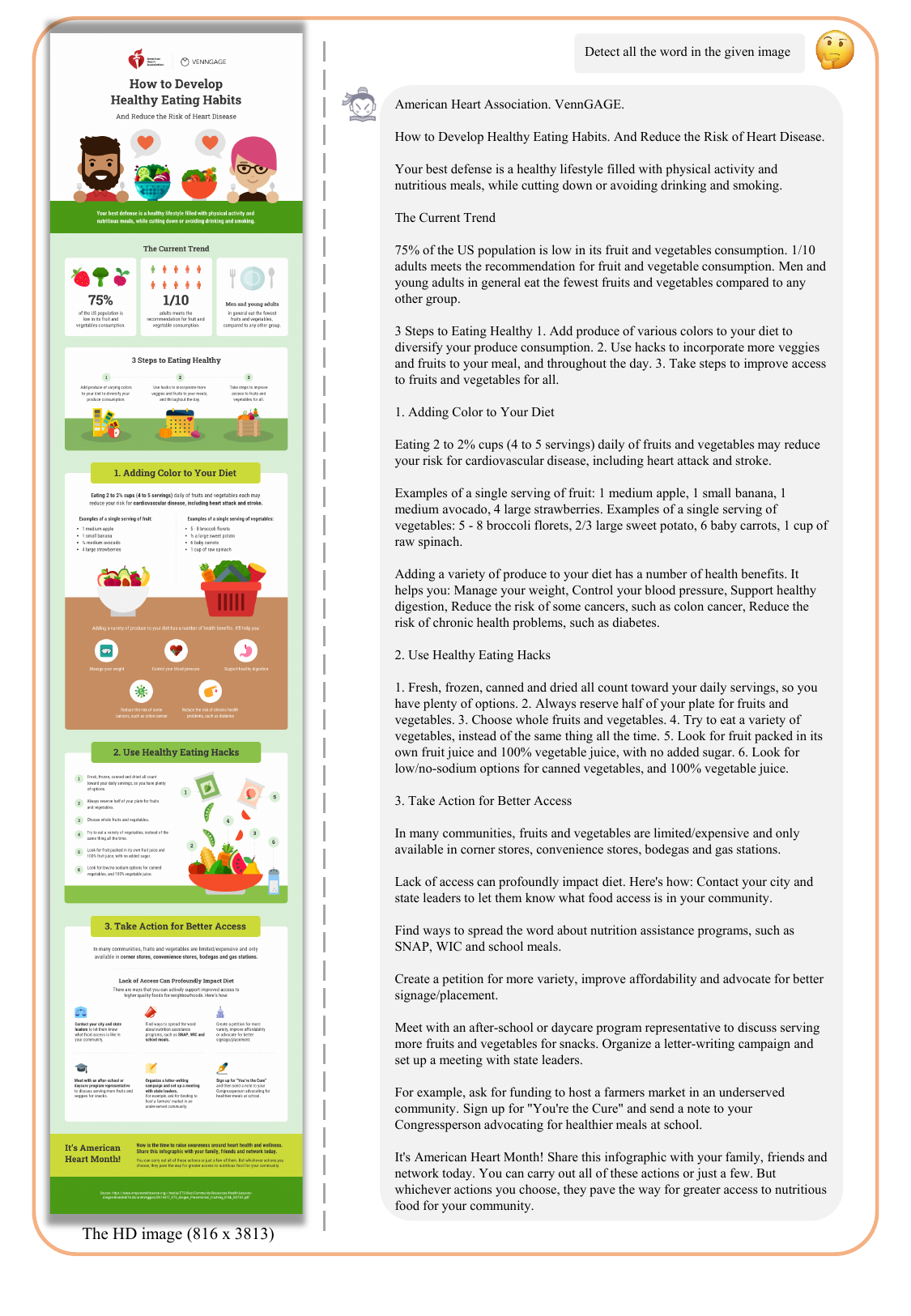}
    \setlength{\abovecaptionskip}{0mm}
    \captionof{figure}{\small
        \textbf{Chat with InternLM-XComposer2-4KHD}. Please refer to the \href{https://s3.amazonaws.com/thumbnails.venngage.com/template/a204bd5c-7e6b-4b96-afd2-92c10fb14040.png}{link} for better visualization.
	}
	\label{fig:teaser_6}
\end{center}
}]

% \renewcommand{\thetable}{A\arabic{table}}
% \renewcommand{\thefigure}{A\arabic{figure}}

% \newpage
% \clearpage
% \input{sections/Supplementary}

\end{document}